\documentclass[conference]{IEEEtran}
\IEEEoverridecommandlockouts
\usepackage{cite}
\usepackage{amsmath,amssymb,amsfonts}
\usepackage{algorithmic}
\usepackage{graphicx}
\usepackage{textcomp}

\usepackage{natbib}
\setcitestyle{numbers,square}

\usepackage{CJKutf8}

\usepackage{multirow}
\usepackage[table,xcdraw]{xcolor}
\usepackage{graphicx}
\usepackage{makecell}
\usepackage{booktabs}
\usepackage[justification=centering]{caption}

\begin{document}

\title{Comparative Study on the Discourse Meaning of Chinese and English Media in the Paris Olympics Based on LDA Topic Modeling Technology and LLM Prompt Engineering\\
}

\author{\IEEEauthorblockN{1\textsuperscript{st} Yinglong Yu}
\IEEEauthorblockA{\textit{Communication University of China} \\
Beijing, China \\
yuyingling@cuc.edu.cn}
\and
\IEEEauthorblockN{2\textsuperscript{nd} Zhaopu Yao}
\IEEEauthorblockA{\textit{Communication University of China} \\
Beijing, China \\
yaozp@cuc.edu.cn}
\and
\IEEEauthorblockN{3\textsuperscript{rd} Fang Yuan}
\IEEEauthorblockA{\textit{Communication University of China} \\
Beijing, China \\
yuanfang@cuc.edu.cn}
}

\maketitle

\begin{abstract}
This study analyzes Chinese and English media reports on the Paris Olympics using topic modeling, Large Language Model (LLM) prompt engineering, and corpus phraseology methods to explore similarities and differences in discourse construction and attitudinal meanings. Common topics include the opening ceremony, athlete performance, and sponsorship brands. Chinese media focus on specific sports, sports spirit, doping controversies, and new technologies, while English media focus on female athletes, medal wins, and eligibility controversies. Chinese reports show more frequent prepositional co-occurrences and positive semantic prosody in describing the opening ceremony and sports spirit. English reports exhibit positive semantic prosody when covering female athletes but negative prosody in predicting opening ceremony reactions and discussing women's boxing controversies.
\end{abstract}

\begin{IEEEkeywords}
Extended Unit of Meaning, LDA, Large Language Models, Prompt Engineering
\end{IEEEkeywords}

\section{Introduction}
The Paris Olympics, held from July 26 to August 11, 2024, marked France's return to hosting the Summer Games after a 100-year gap. As a global sporting event and ceremonial medium, the Olympics have significant cultural, political, and economic impact, attracting intense media attention worldwide. Media reports not only document the events but also reflect the cultural perspectives and values of the reporting countries, shaping global perceptions of the Olympic spirit and the host nation. Given the complex international context and regional conflicts, how will English-language media frame their coverage to influence public perception, and what aspects will Chinese media highlight, especially in light of the 60th anniversary of diplomatic relations between China and France? What characteristic discourses will emerge, and what attitudes will they convey? These questions warrant exploration.

Existing research on Olympics media coverage primarily focuses on single languages\cite{WTXB202203003}\cite{YLYY202301001}, with limited comparative studies from a multilingual perspective. To fill this gap, this study analyzes news reports on the ``Paris Olympics" from mainstream Chinese and English media using topic modeling and Large Language Model (LLM) prompt engineering. Drawing on Sinclair's theory of extended meaning units\cite{sinclair2004trust}, we analyze reports to reveal similarities and differences in discourse construction and attitudinal meanings in both Chinese and English media coverage at both macro and micro levels.

\section{Related Work}
\subsection{Topic Modeling}

Topic modeling techniques generally refer to the process of uncovering latent topics within a collection of documents. After preprocessing, these documents are fed into a topic model which, through unsupervised learning, outputs a certain number of topics. Each topic comprises several keywords that are associated with it. Researchers typically infer the meaning of each topic based on these keywords.

Since the late twentieth century, when Latent Semantic Indexing (LSI)\citep{papadimitriou1998latent} and Probabilistic Latent Semantic Analysis (PLSA)\citep{hofmann1999probabilistic} were introduced, researchers have proposed numerous topic models. Among them, the most commonly used today remains LDA (Latent Dirichlet Allocation), a three-layer Bayesian topic model introduced by Blei et al.\cite{blei2003latent}. 

This classic topic model and its variants continue to be utilized by researchers even in recent years. Yu and Xiang\cite{yu2023discovering} used the LDA model to study approximately 170,000 articles related to artificial intelligence research from 1990 to 2021, analyzing the distinctive characteristics of AI research across different countries. Kukreja et al.\cite{kukreja2023recent} employed the LDA model to examine 325 papers on mathematical expression recognition from 1967 to 2021, identifying the latest trends in this field. Watanabe and Baturo\cite{watanabe2024seeded} improved upon the LDA model by integrating seed sequences, enhancing the model's effectiveness in topic classification. 

\subsection{Large Language Models}

Since the emergence of models like GPT-3\citep{brown2020language}, LLAMA\citep{touvron2023llama}, and Qwen\citep{bai2023qwen}, natural language processing has entered the era of ``large language models" (LLMs). Researchers use specific prompt words to adapt LLMs to downstream tasks. Prompt engineering involves designing and optimizing prompts to ensure accurate and coherent output.

Early prompt engineering used concise templates. For example, Yin et al.\cite{yin2019benchmarking} defined a template as ``The topic of this document is [Z]" for text classification. Kojima et al.\cite{kojima2022large} improved reasoning skills through few-shot learning with a ``Let’s think step by step" template. This step-by-step thinking, termed Chain-of-Thought (CoT) by Wei et al.\cite{wei2022chain}, enhances LLM problem-solving abilities. Wu et al.\cite{wu2024quartet} used CoT to create a four-step classification framework (QLFR) for more efficient text classification. Zhang et al.\cite{zhang2024pushing} applied CoT to data samples, improving classification accuracy by using training sample distributions as additional prompts.

\section{Research Design}

\subsection{Research Question}

The research questions addressed in this study are as follows:
\begin{itemize}
    \item [1)] 
    What are the main topics that Chinese and English mainstream media focus on when reporting on the Paris Olympics? How do these topics manifest specifically in the coverage?
    \item [2)] 
    How do the extended meaning units of keywords differ in specific topics between Chinese and English media reports? What differences in media attitudes do these reflect?

\end{itemize}

\subsection{Corpus Source}

\begin{figure*}[t]
\centering
  \includegraphics[width=0.6\linewidth]{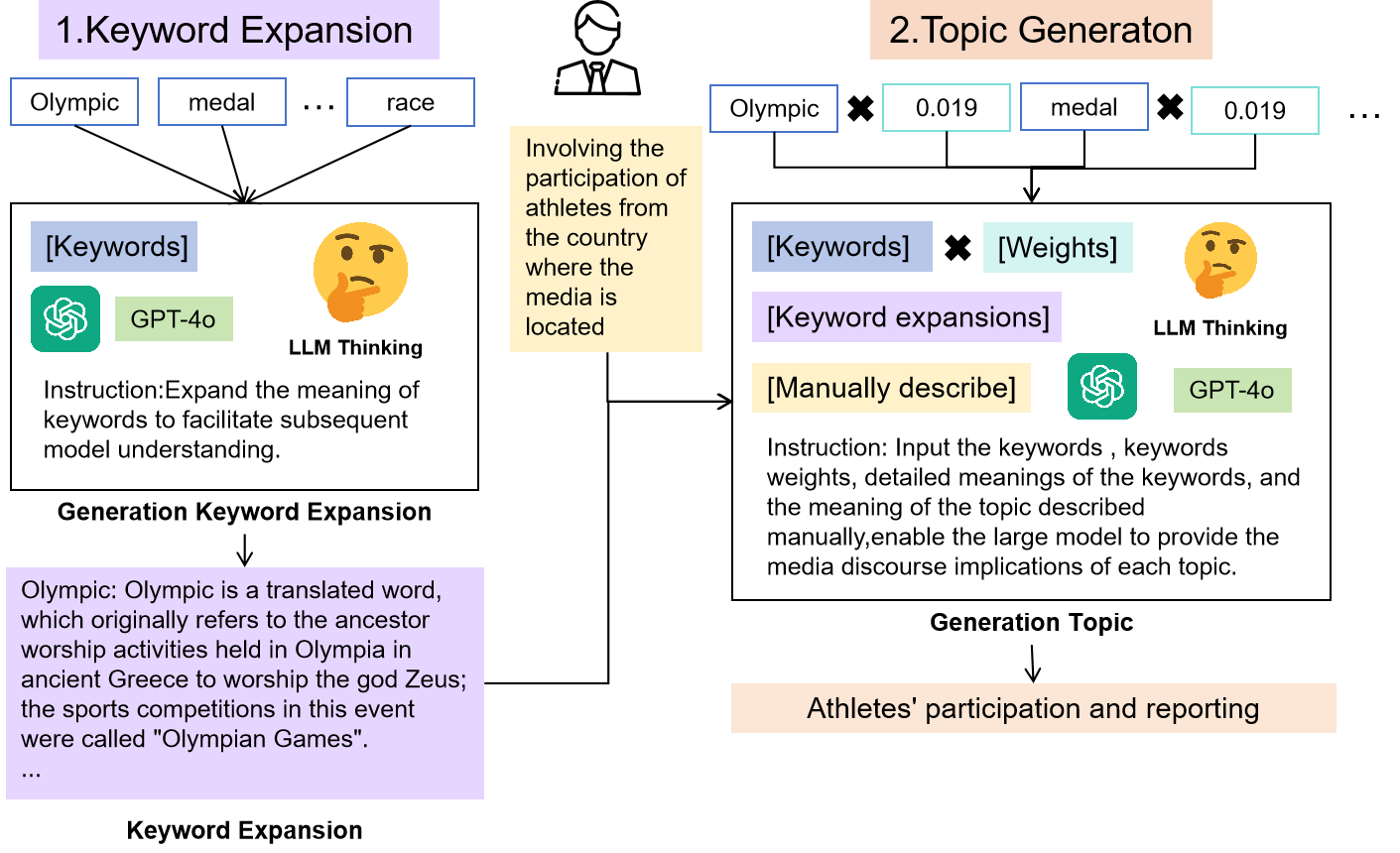}
  \caption{Prompt engineering framework.}
  \label{fig3}
\end{figure*}

In this study, we constructed a corpus of media news reports in both Chinese and English covering the ``Paris Olympics." The materials were extracted using the Wisdom News Search Research Database for Chinese media and the Factiva Dow Jones News and Business Library for English media. We selected ten top-tier Chinese media outlets, including the People's Daily and Xinhua News Agency, along with ten well-known English media outlets, such as The Washington Post and The New York Times. The specific list of media outlets and the rationale for their selection are detailed in Appendix~\ref{appendixA}. The time frame covered spans from half a month before the opening ceremony to half a month after the opening ceremony of the Paris Olympics (July 12 to August 12, 2024). Search keywords included 
\begin{CJK}{UTF8}{gbsn}
``奥运会(Olympics Games)," ``奥运(Olympic)," ``奥林匹克(Olympiad)," ``Olympic Games," ``Olympic," and ``Olympics."
\end{CJK} 
News reports with fewer than two keyword occurrences were excluded. After cleaning and analysis, the final dataset consisted of 715 Chinese media reports and 499 English media reports, totaling 863,572 tokens.

\subsection{Prompt Engineering Framework Design}

LDA topic model results, represented by word groups, can include incoherent or unrelated words, making interpretation difficult. To address this, we used large language model prompt engineering to generate coherent topic descriptions.

First, collect the top $k$ keywords and their corresponding topic weights for each topic, and manually describe the ``probable" general content of each topic. If the topic keywords are too abstract or incomprehensible, they should be skipped without description. Second, have a large language model review its knowledge base to generate detailed meanings for the top $k$ keywords of each topic, which will facilitate further understanding by the model. Finally, input the keywords , keywords weights, detailed meanings of the keywords, and the meaning of the topic described manually. This will enable the large model to provide the media discourse implications of each topic.As shown in the Figure \ref{fig3}.

The specific prompt engineering formula is as follows:
\begin{equation}
  \label{eq:example1}
  \mathrm{S}_{\mathrm{ij}}=\mathrm{F}_{\mathrm{retrieve}}\left(\sum_{\mathrm{i}=1}^{k} \mathrm{~W}_{\mathrm{ij}}, \mathrm{I}_{1}\right)
\end{equation}
\begin{equation}
  \label{eq:example2}
  \mathrm{T_i=F_{topic}(\sum_{j=1}^{k}(W_{ij}\times V_{ij}+S_{ij})+P_i,I_2)}
\end{equation}

Where $\mathrm{I}_{i}$ represents the $i^{th}$ prompt instruction set, $\mathrm{S}_{\mathrm{ij}}$ denotes the detailed meaning of the $j^{th}$ keyword for the $i^{th}$ topic, $\mathrm{F}_{\mathrm{retrieve}}$ is the function enabling the model to recall relevant information from its internal knowledge base, $W_{ij}$ represents the $j^{th}$ keyword for the $i^{th}$ topic, $\mathrm{F}_{\mathrm{topic}}$ is the function prompting the model to generate the media discourse implications of the topic, $V_{ij}$ denotes the weight of the $j^{th}$ keyword for the $i^{th}$  topic, and $P_i$ represents the manually described general content of the $i^{th}$ topic.

\subsection{Research Steps}

This study followed these steps for analysis:

First, we preprocessed the corpus using a large language model to remove irrelevant text, eliminate stopwords and garbled characters, and perform lemmatization on the remaining tokens.

\begin{figure*}[t]
\centering
  \begin{minipage}{0.45\textwidth}
    \centering
    \includegraphics[width=\textwidth]{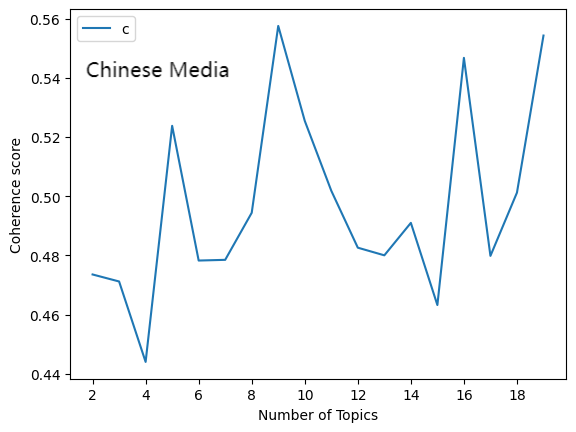}
  \end{minipage}
  \hfill
  \begin{minipage}{0.45\textwidth}
    \centering
    \includegraphics[width=\textwidth]{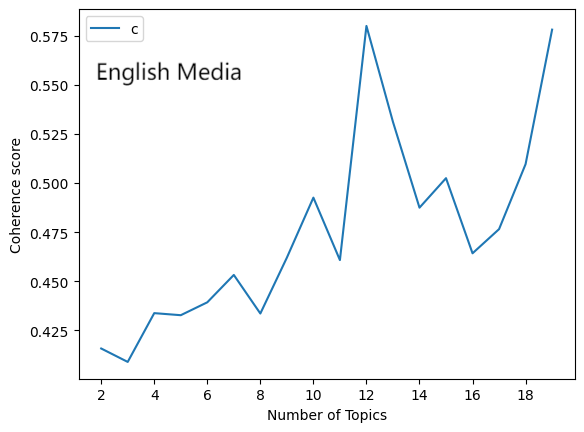}
  \end{minipage}
  \caption{Trend of consistency scores between Chinese and English media coverage Topics.}
  \label{fig1}
\end{figure*}

Second, we applied LDA models separately to the Chinese and English media report corpora. Through a series of experiments with topic numbers ranging from 2 to 20, we determined the optimal number of topics using ``topic coherence" as an evaluation metric. The optimal number of topics was 9 for Chinese media reports and 12 for English media reports, with 10 keywords per topic. The experimental results are shown in Figure~\ref{fig1}.

Next, select high-probability-weight and high-frequency keywords within each topic as node words. Statistically analyze the frequent collocates and co-occurrence patterns of these node words, and determine their semantic trends and semantic prosody by examining instances of their use in broader contexts. During collocation analysis, consider that different forms of the same lemma might have different collocates, usage, and meanings; therefore, analysis should be based on word forms rather than lemma items themselves\citep{sinclair2004trust}.

Finally, combine the collocates, semantic prosody, and lexical indices of the node words to analyze the construction of discourses around the Paris Olympics and the attitudes and meanings conveyed by the media.

\section{Differences and Similarities in Topic Distribution of Chinese and English Media Reports on the Paris Olympics}

In critical discourse analysis, the optimal number of topics is determined not only statistically but also based on their relevance to the research questions and practical significance. After manual evaluation and merging of keywords, we selected 7 topics for the Chinese media and 6 for the English media. These topics are clearly differentiated and form coherent semantic clusters without overlap.

The LDA topic modeling results are summarized in Tables \ref{tab1} and \ref{tab2}, listing 10 keywords per topic with their relevance weights. Higher weights indicate a greater likelihood of the keyword appearing within the topic.

\subsection{Distribution of Topics in Chinese Media Reports on the Paris Olympics}

Based on the thematic results shown in Table \ref{tab1}, the coverage of the Paris Olympics by Chinese media can be summarized into the following seven categories:
\begin{CJK}{UTF8}{gbsn}
(1) Outstanding areas of achievement, including terms such as 郑钦文(Zheng Qinwen), 王楚钦(Wang Chuqin), 樊振东(Fan Zhendong), 马龙(Ma Long), etc., reflect the Chinese media's focus on areas where Chinese athletes excel, such as tennis and table tennis.
(2) The opening ceremony of the Paris Olympicss, including terms such as 法国(France), 举办(host), 开幕式(opening ceremony), and 塞纳河(Seine River).
(3) Athlete performance, including terms such as 比赛(competition), 金牌(gold medal), 冠军(champion), and 中国(China).
(4) Sportsmanship, including terms such as 体育(sports), 文化(culture), 精神(spirit), and 发展(development).
(5) New Olympic technologies, including terms such as 技术(technology), 科技(science), 智能(smart), and 自动(automation).
(6) Olympic sponsorship brands, such as 品牌(brand), 成长(growth), 市场(market) and 商业(enterprise).
(7) Controversy over Doping Testing, including terms such as 兴奋剂(doping), 美国(United States), 检测(testing), and 污染(contamination).
\end{CJK}

\subsection{Distribution of Topics in English Media Reports on the Paris Olympics}

Based on the thematic results shown in Table \ref{tab2}, the coverage of the Paris Olympics by English media can be summarized into the following six categories:
(1) Stories about popular athletes at the Paris Olympics, including terms such as Maher, Chiles, Kim, and TikTok.
(2) Olympic sponsorship brands, including terms such as Nike, Lululemon, and shopping.
(3) The opening ceremonies of the Paris Olympics, including terms such as Paris, Ceremony, and Seine.
(4) Athlete performance, including terms such as medal, gold, won, and athletes.
(5) The performance of female athletes, including terms such as women, Biles, and champion.
(6) Coverage of participation rights and controversies, including terms such as Russian, Ukraine, and Khelif.

\subsection{Similarities and differences in Topic Distribution of Chinese and English Media Reports on the Paris Olympics}

Comparing the results in Table \ref{tab1} and Table \ref{tab2} allows us to analyze the similarities and differences in the distribution of topics in Chinese and English media reports on the Paris Olympics as follows:

Both Chinese and English media reports focus on similar topics such as the ``the opening Ceremony of the Paris Olympics," ``Athlete Performance," and ``Olympic Sponsorship Brands."

Distinctive topics include:

(1) Chinese Media Reports: ``Outstanding Performance Areas" (e.g., tennis and table tennis), ``Spirit of Sports," ``Doping Test Controversies," and ``New Olympic Technologies."

(2) English Media Reports: ``Paris Olympics Influencer Athlete Stories" (e.g., Ilona Maher in rugby), ``Performance of Female Athletes" (e.g., swimming), and ``Eligibility and Controversies."

Commonalities exist within these distinctive topics, such as focusing on areas where athletes excel and the performance of female athletes.

These similarities and differences reflect a shared interest in the opening ceremony, events, and sponsorship brands. Chinese media emphasize the Olympic spirit and new technologies, while addressing doping controversies. English media highlight gender equality and international issues, including eligibility and controversies involving transgender athletes and athletes from countries in international disputes.

\section{Analysis of Extended Meaning Units for Keywords in Topics of Chinese and English Media Reports}

\begin{CJK}{UTF8}{gbsn}
For the aforementioned topics in Chinese and English media reports, this study selects the three most prevalent topics to analyze the extended meaning units of keywords. One keyword with high thematic relevance and weight is chosen for each topic. In the Chinese reports, the study selects the topics ``Athlete Performance," ``The opening Ceremony of the Paris Olympics," and ``Sportsmanship," with the respective keywords being ``奥运会(Olympics)," ``开幕式(opening ceremony)," and ``体育(sports)." In the English reports, the study selects the topics ``Athlete Performance," ``The opening ceremonies of the Paris Olympics," and ``The performance of female athletes," with the respective keywords being ``medal," ``opening ceremony," and ``women."
\end{CJK}

\subsection{Analysis of Extended Meaning Units for Keywords in Topics of Chinese Media Report}

\begin{CJK}{UTF8}{gbsn}
\subsubsection{``奥运会(Olympic Game)"}
\end{CJK}

\begin{CJK}{UTF8}{gbsn}
The statistical results show that the term ``奥运会(Olympics)" appears 3,413 times under the topic of ``Athlete Performance." When ``奥运会(Olympics)" is used as a node word, it frequently co-occurs with the pattern ``Preposition + Modifier + 奥运会(Olympics)," which appears 587 times. The preposition observed is ``在(at)," and modifiers include ``巴黎(Paris)," ``东京(Tokyo)," ``本届(this edition)," and ``夏季(Summer)." Based on the semantic categories of the modifiers, they can be divided into three groups: the first group indicates the location or reference to the current Olympics; the second group refers to previous host cities; and the third group specifies the time of the Olympics. Specific examples of each category are shown in the Table \ref{tab3}:

As can be seen from the above table, the pattern ``Preposition (在) + Modifier + 奥运会(Olympics)" is primarily used to introduce the performance of the Chinese Olympic delegation at the current Olympics, previous Olympics, and since participating in the Summer Olympics. This pattern serves an objective descriptive function without a clear semantic prosody tendency.
\end{CJK}

%表3
% \begin{CJK}{UTF8}{gbsn}
% % Table generated by Excel2LaTeX from sheet 'Sheet3'
% \begin{table*}[htbp]
%   \centering
%   \resizebox{\textwidth}{!}{
%     \begin{tabular}{cccl}
%     \toprule
%     \textbf{Semantic Category} & {\textbf{Frequency}} & \textbf{Example of Modifier Words} & \multicolumn{1}{c}{\textbf{Example Sentence}} \\
%     \midrule
%     \multirow{2}[1]{*}{\makecell[c]{Olympic Games host city \\or refer to the current Olympic Games} } & \multirow{2}[1]{*}{441} & \multirow{2}[1]{*}{\makecell[c]{巴黎(Paris) (347), \\本届(current edition) (94)}} & \makecell[c]{(1)据卡塔尔半岛电视台网站报道,\\有多达10500名运动员将在巴黎奥运会上竞技。\\(According to the website of Al Jazeera in Qatar, \\up to 10500 athletes will compete at the Paris Olympics.)} \\
%     \multicolumn{1}{r}{} &       & \multicolumn{1}{r}{} & \makecell[c]{(2)……这也是中国代表团在本届\\奥运会获得的第35枚金牌。\\(This is also the 35th gold medal won by \\the Chinese delegation at this Olympic Games.)} \\
%     Previous Olympic Games host cities & 134   & \makecell[c]{东京(Tokyo) (95), \\里约(Rio) (11), \\洛杉矶(Los Angeles) (10)} & \makecell[c]{(3)中国队在东京奥运会上获得了这个项目的金牌。\\(The Chinese team won the gold medal in \\this event at the Tokyo Olympics.)} \\
%     Indicate the time of the Olympic Games & 12    & 夏季(summer) (12) & \makecell[c]{(4)……这也是中国体育代表团在\\夏季奥运会上的第300金。\\(This is also the 300th gold medal for the Chinese \\sports delegation at the Summer Olympics.)} \\
%     \bottomrule
%     \end{tabular}%
%     }
%     \caption{Semantic classes and specific use cases of modifiers in the pattern of ``Prep+Modifier+Olympics."}
%   \label{tab3}%
% \end{table*}%
% \end{CJK}
\begin{CJK}{UTF8}{gbsn}
% Table generated by Excel2LaTeX from sheet 'Sheet3'
\begin{table*}[htbp]
  \caption{Semantic classes and specific use cases of modifiers in the pattern of ``Prep+Modifier+Olympics."}
  \begin{center}
    \renewcommand{\arraystretch}{1.5}
    \resizebox{\textwidth}{!}{
    \begin{tabular}{|c|c|c|l|}
    \hline
    \textbf{Semantic Category} & \textbf{Frequency} & \textbf{Example of Modifier Words} & \multicolumn{1}{c|}{\textbf{Example Sentence}} \\
    \hline
    \multirow{4}{*}{\makecell[c]{Olympic Games host city \\or refer to the current Olympic Games} } & \multirow{4}{*}{441} & \multirow{4}{*}{\makecell[c]{巴黎(Paris) (347), \\本届(current edition) (94)}} & \makecell[c]{(1)据卡塔尔半岛电视台网站报道,\\有多达10500名运动员将在巴黎奥运会上竞技。\\(According to the website of Al Jazeera in Qatar, \\up to 10500 athletes will compete at the Paris Olympics.)} \\
    \cline{4-4}
    & & & \makecell[c]{(2)……这也是中国代表团在本届\\奥运会获得的第35枚金牌。\\(This is also the 35th gold medal won by \\the Chinese delegation at this Olympic Games.)} \\
    \hline
    Previous Olympic Games host cities & 134 & \makecell[c]{东京(Tokyo) (95), \\里约(Rio) (11), \\洛杉矶(Los Angeles) (10)} & \makecell[c]{(3)中国队在东京奥运会上获得了这个项目的金牌。\\(The Chinese team won the gold medal in \\this event at the Tokyo Olympics.)} \\
    \hline
    Indicate the time of the Olympic Games & 12 & 夏季(summer) (12) & \makecell[c]{(4)……这也是中国体育代表团在\\夏季奥运会上的第300金。\\(This is also the 300th gold medal for the Chinese \\sports delegation at the Summer Olympics.)} \\
    \hline
    \end{tabular}%
    }
  \end{center}
  \label{tab3}%
\end{table*}%
\end{CJK}

\begin{CJK}{UTF8}{gbsn}
\subsubsection{``开幕式(Opening Ceremony)"}
\end{CJK}
\begin{CJK}{UTF8}{gbsn}
Under the topic of ``the opening Ceremony of the Paris Olympics," the term ``开幕式(opening ceremony)" appears 238 times. When ``opening ceremony" is used as a node word, the two most common co-occurrence patterns are ``在(at) + 开幕式(opening ceremony) + 上(on) + VP" and ``开幕式(opening ceremony) + PP," with the former occurring 30 times and the latter 21 times. Further examination of the internal components of these two patterns reveals that, from a semantic class perspective, the verbal component in ``在(at) + 开幕式(opening ceremony) + 上(on) + VP" can be divided into two categories: one representing segments of the opening ceremony, and the other representing audiovisual effects. The prepositional phrase in ``开幕式(opening ceremony) + PP" is used to indicate the location of the opening ceremony. Specific examples are shown in the Table \ref{tab4}:

From a semantic prosody perspective, ``VP" in ``在(at) + 开幕式(opening ceremony) + 上(on) + VP" serves an objective function when referring to ceremony segments, and conveys a positive attitude when referring to audiovisual effects. Objectively stated facts occur 28 times, while positive semantic prosodies occur 2 times. ``Opening ceremony + PP" states the location objectively, without a clear semantic prosody tendency.
\end{CJK}

\begin{CJK}{UTF8}{gbsn}
\subsubsection{``体育(Sports)"}
\end{CJK}

\begin{CJK}{UTF8}{gbsn}
Under the topic of ``Sportsmanship," the term ``sports" appears 1,114 times. The sequence ``中国体育(Chinese sports)" occurs 178 times, making it a significant node word. We analyze its co-occurrence patterns: ``中国体育(Chinese sports) + (N) + VP" (55)\footnote{The numbers in parentheses after the form and example words in this article indicate frequency} and ``V + 中国体育(Chinese sports)" (46).

Without a noun following ``Chinese sports," the pattern ``Chinese sports + VP" (10) refers to achievements and future development. For example:

(8) 自1984年之后，中国体育也完成了诸多“第一”：第一次举办亚运会、第一次举办奥运会、第一次位列奥运金牌榜第一、第一次设立全民健身日。(Since 1984, Chinese sports have accomplished numerous ``firsts": hosting the Asian Games for the first time, hosting the Olympics for the first time, ranking first on the Olympic gold medal list for the first time, and establishing National Fitness Day for the first time.)

(9) 面对下一届洛杉矶奥运会,中国体育勇于迎接更为艰巨的挑战。(Facing the next Los Angeles Olympics, Chinese sports are ready to meet even more challenging tasks.)

These examples show a positive semantic prosody.

With a noun following ``中国体育(Chinese sports)," the pattern ``Chinese sports + N + VP" (45) can be categorized into three groups: people (22), industry (11), and endeavors (3). Examples include:

(10) 在本届奥运会上,中国体育健儿用坚持不懈的拼搏和追赶不断刷新纪录、创造奇迹。(At this Olympic Games, Chinese athletes demonstrated perseverance and relentless pursuit, constantly breaking records and creating miracles.)

(11) 中国体育健儿厉兵秣马,蓄势待发,将再写辉煌。(Chinese athletes are preparing diligently and poised to write new glories.)

(12) 乒乓球、足球、篮球、排球等体育用品的订单量猛增，中国体育用品掀起出口热潮。(Orders for table tennis, soccer, basketball, volleyball, and other sports equipment surged, sparking an export boom in Chinese sports equipment.)

(13) 新时代以来，中国体育事业迎来腾飞。(Since the new era, the Chinese sports endeavor has taken off.)

These examples also convey a positive semantic prosody.

\end{CJK}

%表4
% \begin{CJK}{UTF8}{gbsn}
% % Table generated by Excel2LaTeX from sheet 'Sheet1'
% \begin{table*}[htbp]
%   \centering
%   \resizebox{\textwidth}{!}{
%     \begin{tabular}{cccl}
%     \toprule
%     \textbf{Pattern} & \textbf{Semantic Category} & \textbf{Frequency} & \multicolumn{1}{c}{\textbf{Example Sentence}} \\
%     \midrule
%     \multirow{2}[1]{*}{\makecell[c]{(在(In)) + 开幕式(Opening Ceremony)\\ + 上(On) + VP}} & Opening Ceremony Segment & 28    & \makecell[c]{(5)国际奥委会主席巴赫在开幕式上致辞。\\(International Olympic Committee President Bach\\ delivered a speech at the opening ceremony.) }\\
%           & Audiovisual Effect & 2     & \makecell[c]{(6)塞纳河上有不少桥，在开幕式上让人过目不忘。\\(There are many bridges on the Seine River,\\ which are unforgettable at the opening ceremony.) }\\
%     开幕式(Opening Ceremony) + PP & Opening Ceremony Location & 21    & \makecell[c]{(7)第33届夏季奥运会开幕式在塞纳河上拉开帷幕。\\(The opening ceremony of the 33rd \\Summer Olympics kicked off on the Seine River.)} \\
%     \bottomrule
%     \end{tabular}%
%     }
%     \caption{The high-frequency co-occurrence pattern of ``Opening Ceremony"\\ consists of semantic classes and use cases.}
%   \label{tab4}%
% \end{table*}%
% \end{CJK}
\begin{CJK}{UTF8}{gbsn}
\begin{table*}[htbp]
  \caption{The high-frequency co-occurrence pattern of ``Opening Ceremony'' consists of semantic classes and use cases.}
  \begin{center}
    \renewcommand{\arraystretch}{1.5}
    \resizebox{\textwidth}{!}{
    \begin{tabular}{|c|c|c|l|}
    \hline
    \textbf{Pattern} & \textbf{Semantic Category} & \textbf{Frequency} & \multicolumn{1}{c|}{\textbf{Example Sentence}} \\
    \hline
    \multirow{4}{*}{\makecell[c]{(在(In)) + 开幕式(Opening Ceremony)\\ + 上(On) + VP}} & \multirow{2}{*}{Opening Ceremony Segment} & \multirow{2}{*}{28} & \makecell[c]{(5)国际奥委会主席巴赫在开幕式上致辞。\\(International Olympic Committee President Bach\\ delivered a speech at the opening ceremony.) }\\
    \cline{4-4}
    & \multirow{2}{*}{Audiovisual Effect} & \multirow{2}{*}{2} & \makecell[c]{(6)塞纳河上有不少桥，在开幕式上让人过目不忘。\\(There are many bridges on the Seine River,\\ which are unforgettable at the opening ceremony.) }\\
    \hline
    开幕式(Opening Ceremony) + PP & Opening Ceremony Location & 21 & \makecell[c]{(7)第33届夏季奥运会开幕式在塞纳河上拉开帷幕。\\(The opening ceremony of the 33rd \\Summer Olympics kicked off on the Seine River.)} \\
    \hline
    \end{tabular}%
    }
  \end{center}
  \label{tab4}%
\end{table*}%
\end{CJK}

\subsection{Analysis of Extended Meaning Units for Keywords in  Topics of English Media Report}

\subsubsection{``medal"}

Under the topic of ``Athlete Performance," the term ``medal" appears 1,602 times. This study focuses on the node word phrase ``(the) gold medal" and its co-occurrence patterns. The two most frequent patterns are ``V + (Modifier) + (the) gold medal" (118) and ``Prep + (the) gold medal + N" (58).

In ``V + (Modifier) + (the) gold medal," from a semantic category perspective, the verbs can be classified into four types: the first type is acquisition (38), such as ``won," ``received," ``secured," and ``got"; the second type is sharing (8), such as ``shared"; the third type is loss (7), such as ``missed out" and ``removed from"; and the fourth type is pursuit (7), such as ``aiming for" and ``looking for". ``V + (Modifier) + (the) gold medal" includes cases with and without modifiers. When modifiers are present, they mainly consist of two types: the first type is ordinal numbers (101), such as ``first," ``third," ``second," and ``fourth"; the second type is activity names (17), such as ``Olympic" and ``equestrian". Examples include:

(14)He ……over Fiji in the Rugby Sevens final and won France's first gold medal of its home Olympic Games.

(15)Paris and London share the gold medal.

(16)And it is an awful time by which to miss out on an Olympic gold medal after four years of training.

(17)Tom Daley, 30, the Team GB diver aiming for a second gold medal.

Analysis shows that the pattern tends to convey a positive semantic prosody in most cases, with a negative prosody in a few instances (e.g., missing out on a gold medal).

\subsubsection{``opening ceremony''}

Under the topic of ``the opening Ceremony of the Paris Olympics,'' the term ``(the) opening ceremony'' appears 245 times. Examination reveals that it frequently co-occurs with the pattern ``VP + Prep + (the) opening ceremony,'' which occurs 97 times. From a semantic category perspective, the Verb Phrases (VPs) can be classified into four types: the first type is performance or specific ceremony-related (50); the second type is preparation-related (12); the third type is situation-related (5); and the fourth type is unexpected situation-related (5).

Observation of the index lines for the various VPs reveals that performance or specific ceremony-related VPs commonly co-occur with the prepositions ``during,'' ``at,'' and ``in''; preparation-related VPs commonly co-occur with the prepositions ``ahead of,'' ``before,'' ``prior,'' and ``for''; situation-related VPs commonly co-occur with the preposition ``after''; and unexpected situation-related VPs commonly co-occur with the preposition ``during.'' Examples include:

(18) Lady Gaga rehearses prior to the opening ceremony of the Olympic Games.

(19) A torchbearer runs during the opening ceremony in Paris on July 26.

(20) ...as criticism continues to grow after the opening ceremony made a mockery of the Last Supper with a recreation that featured performers in drag.

(21) He lost his wedding ring in the Seine during the opening ceremony.

In examples (18)-(21), verb phrases co-occur with prepositions and the node word to describe events during and around the opening ceremony:(1) "Lady Gaga rehearses" (18) describes preparations before the ceremony.(2) "A torchbearer runs" (19) depicts a segment of the torch relay during the ceremony.(3) "Criticism continues to grow" (20) predicts increased criticism after the ceremony.(4) "He lost his wedding ring in the Seine" (21) recounts an incident during the ceremony.

Descriptions are generally objective, with some negative semantic prosody in predictions post-ceremony.

\subsubsection{``women''}

Under ``The performance of female athletes'' ``women'' appears 1,253 times, strongly co-occurring with "'s" (957). Thus, we use ``women's'' as the node word. The high-frequency pattern is ``women's + (Modifier) + N,'' where the modifier indicates the form of competition (e.g., ``57 kilogram judo''). Common nouns include ``medley'' (56), ``rugby'' (30), ``basketball'' (30), ``boxing'' (27), ``freestyle'' (26), ``soccer'' (20), ``butterfly'' (22), ``judo'' (19), ``air pistol'' (17), ``rowing'' (14), ``hockey'' (13), ``sprint'' (13), ``canoe'' (12), ``backstroke'' (12), ``volleyball'' (10), ``marathon'' (10), ``football'' (10), and ``weightlifting'' (10). Focus areas are swimming (116), ball games (113), combat sports (46), boat sports (26), and running (23). Verb phrases often precede ``women's + (Modifier) + N'' and can be positive, neutral, or negative, influencing semantic prosody. See Table \ref{tab5} for examples.

%表5
% \begin{table*}[htbp]
% \resizebox{\textwidth}{!}{%
% \begin{tabular}{ccl}
% \hline
% \textbf{\makecell[c]{Left-hand Verb \\ Component Category}} &
%   \multicolumn{1}{c}{\textbf{Use Case}} &
%   \multicolumn{1}{c}{\textbf{Example Sentence}} \\ \hline
% Positive expression &
%   \makecell[c]{``win/won,win/won(the)gold''(53),\\``coming out on top''(9),\\``broke record''(7),dominated(3),\\``set a new record''(1)} &
%   \makecell[c]{(22)McIntosh won Canada's second gold medal, coming out on top \\in the women's 400-metre individual medley with a time of 4:27.71.} \\
% Neutral expression &
%   \makecell[c]{``compete''(8)} &
%   \makecell[c]{(23)Athletes compete in the women's 10,000m finalat the at \\ Stade de France in Saint-Denis, north of Paris, on Friday.} \\
% Negative expression &
%   \makecell[c]{``disqualified''(8),\\``the storm that has erupted over''(2),\\``complain about the justice ''(2)} &
%   \makecell[c]{(24)The following day we had our own reasons to \\ complain about the justice of women's boxing in the Olympics.} \\ \hline
% \end{tabular}%
% }
% \caption{``women’s+(Modifier)+N'' Left co-occurrence of verb elements and example sentences.}
% \label{tab5}
% \end{table*}
\begin{table*}[htbp]
\caption{``women's+(Modifier)+N'' Left co-occurrence of verb elements and example sentences.}
\begin{center}
\renewcommand{\arraystretch}{1.5}
\resizebox{\textwidth}{!}{%
\begin{tabular}{|c|c|l|}
\hline
\textbf{\makecell[c]{Left-hand Verb \\ Component Category}} &
  \textbf{Use Case} &
  \multicolumn{1}{c|}{\textbf{Example Sentence}} \\ 
\hline
\multirow{2}{*}{Positive expression} &
  \multirow{2}{*}{\makecell[c]{``win/won,win/won(the)gold''(53),\\``coming out on top''(9),\\``broke record''(7),dominated(3),\\``set a new record''(1)}} &
  \multirow{2}{*}{\makecell[c]{(22)McIntosh won Canada's second gold medal, coming out on top \\in the women's 400-metre individual medley with a time of 4:27.71.}} \\
& & \\
\hline
\multirow{2}{*}{Neutral expression} &
  \multirow{2}{*}{\makecell[c]{``compete''(8)}} &
  \multirow{2}{*}{\makecell[c]{(23)Athletes compete in the women's 10,000m finalat the at \\ Stade de France in Saint-Denis, north of Paris, on Friday.}} \\
& & \\
\hline
\multirow{2}{*}{Negative expression} &
  \multirow{2}{*}{\makecell[c]{``disqualified''(8),\\``the storm that has erupted over''(2),\\``complain about the justice ''(2)}} &
  \multirow{2}{*}{\makecell[c]{(24)The following day we had our own reasons to \\ complain about the justice of women's boxing in the Olympics.}} \\ 
& & \\
\hline
\end{tabular}%
}
\end{center}
\label{tab5}
\end{table*}

In example (22), ``coming out on top'' co-occurs with ``women's 400-metre individual medley,'' expressing affirmation of the athlete's achievement in the women's 400-metre individual medley race. In example (23), ``compete'' co-occurs with ``women's 10,000m,'' providing an objective statement of the athlete's participation. In example (24), ``complain about the justice'' co-occurs with ``women’s boxing,'' expressing dissatisfaction with perceived unfairness in the women’s boxing competition in a negative manner. Overall, considering the sentiment of the verb phrases to the left, their frequency, and the extended context, when positive verb phrases co-occur with ``women’s + (Modifier) + N,'' they mostly exhibit a positive and favorable semantic prosody. However, in a few cases, they exhibit a negative semantic prosody, expressing dissatisfaction with unfairness in the competition.

\subsection{Analysis of Differences in Extended Meaning Units of Keywords in Topics of Chinese and English Media Reports}

\begin{CJK}{UTF8}{gbsn}
fThis study analyzed the extended meaning units of keywords in Chinese and English media reports on the Paris Olympics. Three topics were selected: ``Athlete Performance,'' ``The Opening Ceremony of the Olympics,'' and two characteristic topics (``Sportsmanship'' for Chinese reports and ``The Performance of Female Athletes'' for English reports).

\textbf{Athlete Performance:}

(1) Chinese Reports: Keyword ``奥运会(Olympics)'' with the pattern ``Prep + 在(in/at) + Modifier + 奥运会(Olympics)'' for objective statements about athletes' performances.
(2) English Reports: Keyword ``medal'' with the pattern ``V + (Modifier) + (the) gold medal'' exhibiting a positive semantic prosody for gold medal winners and a negative prosody for non-gold medalists.

\textbf{The Opening Ceremony of the Olympics:}

(1) Chinese Reports: Keyword ``开幕式(opening ceremony)'' with the patterns ``在(At/in) + 开幕式(opening ceremony) + 上(on) + VP'' and ``开幕式(opening ceremony) + PP'' for objective descriptions, occasionally showing a positive semantic prosody.
(2) English Reports: Keyword ``opening ceremony'' with the pattern ``VP + Prep + (the) opening ceremony'' for objective descriptions, with a negative semantic prosody in predicting reactions to the opening ceremony.

\textbf{Characteristic Topics:}

(1) Chinese Reports: Keyword ``体育(sports)'' with the pattern ``中国体育(Chinese sports) + (N) + VP'' exhibiting a positive semantic prosody for the spirit of Chinese athletes.

(2) English Reports: Keyword ``women'' with the pattern ``women's + (Modifier) + N'' reflecting attention to female athletes, with a positive semantic prosody except in the context of women's boxing, where a negative prosody is observed.

Overall, Chinese reports frequently use prepositions in their patterns, while English reports do not use prepositions as prominently. Both types of reports are generally objective, with occasional positive or negative semantic prosodies.

\end{CJK}

\section{Conclusion}

This study uses Latent Dirichlet Allocation (LDA) and Large Language Model (LLM) techniques to analyze similarities and differences in Chinese and English media coverage of the Paris Olympics. We find that both media types focus on the opening ceremony, events, and sponsorship brands. Distinctive topics include:

(1) Chinese Media: Tennis, table tennis, Olympic spirit, new technologies, and doping controversies.

(2) English Media: Female athlete performance, particularly in swimming and rugby, and eligibility controversies.

\begin{CJK}{UTF8}{gbsn}
In terms of discourse representation, frequent patterns in Chinese media include ``Prep + 在 + Modifier + Olympics'' and ``中国体育 + (N) + VP,'' while English media patterns include ``V + (Modifier) + (the) gold medal'' and ``women’s + (Modifier) + N.'' Chinese reports have more frequent prepositional co-occurrences. Both media exhibit objectivity, with positive prosodies for opening ceremonies and female athlete performances.
\end{CJK}

Methodologically, LDA and LLM techniques enable rapid topic extraction and avoid reference corpus interference, producing objective findings. This approach overcomes limitations in traditional discourse analysis and is suitable for macro-level thematic analysis of large corpora. By examining keyword collocations and semantic prosodies, we conduct micro-level discourse construction analysis. Future work could enhance LDA integration with practical prompt engineering for more nuanced media discourse analysis.

%\begin{thebibliography}{99}
\bibliographystyle{IEEEtran}
\bibliography{main}
%\end{thebibliography}
% \bibitem{b1} G. Eason, B. Noble, and I. N. Sneddon, ``On certain integrals of Lipschitz-Hankel type involving products of Bessel functions,'' Phil. Trans. Roy. Soc. London, vol. A247, pp. 529--551, April 1955.
% \bibitem{b2} J. Clerk Maxwell, A Treatise on Electricity and Magnetism, 3rd ed., vol. 2. Oxford: Clarendon, 1892, pp.68--73.
% \bibitem{b3} I. S. Jacobs and C. P. Bean, ``Fine particles, thin films and exchange anisotropy,'' in Magnetism, vol. III, G. T. Rado and H. Suhl, Eds. New York: Academic, 1963, pp. 271--350.
% \bibitem{b4} K. Elissa, ``Title of paper if known,'' unpublished.
% \bibitem{b5} R. Nicole, ``Title of paper with only first word capitalized,'' J. Name Stand. Abbrev., in press.
% \bibitem{b6} Y. Yorozu, M. Hirano, K. Oka, and Y. Tagawa, ``Electron spectroscopy studies on magneto-optical media and plastic substrate interface,'' IEEE Transl. J. Magn. Japan, vol. 2, pp. 740--741, August 1987 [Digests 9th Annual Conf. Magnetics Japan, p. 301, 1982].
% \bibitem{b7} M. Young, The Technical Writer's Handbook. Mill Valley, CA: University Science, 1989.
% \end{thebibliography}
\vspace{12pt}

\appendices
\section{Source and Reasons for Choosing Corpus Text Media}
\label{appendixA}
For this study, we constructed a corpus of news reports from Chinese and English-language media outlets covering the ``Paris Olympic Games.'' The materials were extracted using the Wisdom News Search Research Database for Chinese media and the Factiva Dow Jones News \& Business Library for English media. Specifically, ten Chinese media outlets were selected: People's Daily, Xinhua News Agency, China Central Television (CCTV), Qiushi Journal, PLA Daily, Guangming Daily, Economic Daily, China Daily, Science and Technology Daily, and the People's Political Consultative Conference Daily. Ten English-language media outlets were also chosen: The Washington Post, The New York Times, Fox News, National Public Radio (NPR), The Wall Street Journal, The Times, Financial Times, The Economist, The Australian, and The Toronto Star.

The reason for selecting these ten Chinese media outlets is that they all rank among the top ten central news organizations listed in the ``List of Internet News Information Sources'' published by the Cyberspace Administration of China. The ten English-language media outlets were chosen because they enjoy high reputations globally and have been commonly used as data sources in previous studies by scholars such as Wang and Wang\cite{ZGKT202406016} and Gao and Xu \cite{XWXZ202005009}. These twenty media outlets not only possess significant authority and wide-ranging influence but also have produced extensive coverage of the ``Paris Olympic Games.''

\section{Results of the Topic Modeling for Chinese and English Media}
\label{appendixB}
Because the table is too wide and long, the table is placed on the next page for the convenience of readers' viewing.

\begin{CJK}{UTF8}{gbsn}
\begin{table*}[h]
  \caption{Topic modeling results for the subcorpus of Chinese media reports.}
  \begin{center}
  \renewcommand{\arraystretch}{1.5}
  \resizebox{\textwidth}{!}{
    \begin{tabular}{|c|c|l|}
    \hline
    \textbf{Topic number} & \textbf{Topic} & \multicolumn{1}{c|}{\textbf{Top 10 Keywords and Their Weights in Terms of Topic Relevance}} \\
    \hline
    \multirow{3}{*}{1} & \multirow{3}{*}{Outstanding areas of achievement} & \multirow{3}{*}{\makecell[c]{0.042*``郑钦文'' + 0.036*``网球'' + 0.021*``樊振东'' + 0.016*``马龙'' \\+ 0.015*``单打'' + 0.011*``李娜'' + 0.009*``王楚钦'' + 0.008*``卡尔松'' \\+ 0.007*``男单'' + 0.007*``两局''}} \\
    & & \\
    & & \\
    \hline
    \multirow{3}{*}{2} & \multirow{3}{*}{The opening ceremony of the Paris Olympics} & \multirow{3}{*}{\makecell[c]{0.034*``巴黎'' + 0.031*``奥运会'' + 0.021*``运动员'' + 0.016*``法国'' \\+ 0.009*``记者'' + 0.009*``举办'' + 0.009*``开幕式'' + 0.008*``塞纳河'' \\+ 0.007*``国际奥委会'' + 0.006*``期间''}} \\
    & & \\
    & & \\
    \hline
    \multirow{3}{*}{3} & \multirow{3}{*}{Athlete performance} & \multirow{3}{*}{\makecell[c]{0.040*``奥运会'' + 0.019*``项目'' + 0.018*``中国'' + 0.018*``比赛'' \\+ 0.016*``运动员'' + 0.015*``奥运'' + 0.015*``巴黎'' + 0.011*``金牌'' \\+  0.009*``冠军'' + 0.008*``选手''}} \\
    & & \\
    & & \\
    \hline
    \multirow{3}{*}{4} & \multirow{3}{*}{Sportsmanship} & \multirow{3}{*}{\makecell[c]{0.028*``体育'' + 0.015*``中国'' + 0.011*``文化'' + 0.011*``更'' \\+ 0.010*``运动'' + 0.010*``精神'' + 0.007*``发展'' + 0.007*``奥运'' \\+ 0.006*``活动'' + 0.005*``新''}} \\
    & & \\
    & & \\
    \hline
    \multirow{3}{*}{5} & \multirow{3}{*}{New Olympic technologies} & \multirow{3}{*}{\makecell[c]{0.016*``奥运会'' + 0.014*``技术'' + 0.014*``巴黎'' + 0.012*``中国'' \\+ 0.011*``科技'' + 0.011*``运动员'' + 0.010*``赛事'' + 0.009*``体育'' \\+ 0.008*``系统'' + 0.007*``提供''}} \\
    & & \\
    & & \\
    \hline
    \multirow{3}{*}{6} & \multirow{3}{*}{Olympic sponsorship brands} & \multirow{3}{*}{\makecell[c]{0.018*``品牌'' + 0.011*``紫色'' + 0.010*``增长'' + 0.009*``市场'' \\+ 0.008*``产品'' + 0.008*``运动'' + 0.007*``企业'' + 0.007*``消费'' \\+ 0.006*``新'' + 0.006*``更''}} \\
    & & \\
    & & \\
    \hline
    \multirow{3}{*}{7} & \multirow{3}{*}{Controversy over Doping Testing} & \multirow{3}{*}{\makecell[c]{0.024*``兴奋剂'' + 0.023*``美国'' + 0.019*``反'' + 0.019*``运动员'' \\+ 0.006*``检测'' + 0.006*``事件'' + 0.005*``污染'' + 0.005*``国际'' \\+ 0.005*``阳性'' + 0.004*``检查''}} \\
    & & \\
    & & \\
    \hline
    \end{tabular}%
    }
  \end{center}
  \label{tab1}%
\end{table*}%
\end{CJK}

\begin{table*}[htbp]
  \caption{Topic modeling results for the subcorpus of English media reports.}
  \begin{center}
  \renewcommand{\arraystretch}{1.5}
  \resizebox{\textwidth}{!}{
    \begin{tabular}{|c|c|l|}
    \hline
    \textbf{Topic number} & \textbf{Topic} & \multicolumn{1}{c|}{\textbf{Top 10 Keywords and Their Weights in Terms of Topic Relevance}} \\
    \hline
    \multirow{3}{*}{1} & \multirow{3}{*}{Stories of Internet celebrity athletes at the Paris Olympics} & \multirow{3}{*}{\makecell[c]{0.008*``Maher'' + 0.008*``Rivera'' + 0.007*``Chiles'' + 0.006*``Kim'' \\+ 0.006*``online'' + 0.005*``TikTok'' + 0.004*``shooter'' + 0.004*``Ilona'' \\+ 0.004*``shooters'' + 0.004*``Hezly''}} \\
    & & \\
    & & \\
    \hline
    \multirow{3}{*}{2} & \multirow{3}{*}{Olympic sponsorship brands} & \multirow{3}{*}{\makecell[c]{0.010*``Nike'' + 0.005*``Lululemon'' + 0.003*``earnings'' + 0.003*``analysts'' \\+ 0.002*``shopping'' + 0.001*``upbeat'' + 0.001*``sexually'' + 0.001*``accepting'' \\+ 0.001*``expects'' + 0.001*``absurdly''}} \\
    & & \\
    & & \\
    \hline
    \multirow{3}{*}{3} & \multirow{3}{*}{The opening ceremonies of the Paris Olympics} & \multirow{3}{*}{\makecell[c]{0.024*``Paris'' + 0.020*``French'' + 0.017*``France'' + 0.015*``ceremony'' \\+ 0.015*``stadium'' + 0.015*``ceremony'' + 0.013*``Games'' + 0.012*``Seine'' \\+ 0.010*``Cruise'' + 0.010*``flag''}} \\
    & & \\
    & & \\
    \hline
    \multirow{3}{*}{4} & \multirow{3}{*}{Athlete performance} & \multirow{3}{*}{\makecell[c]{0.025*``Olympics'' + 0.024*``Games'' + 0.023*``Olympic'' + 0.018*``medal'' \\+ 0.017*``gold'' +  0.016*``Paris'' + 0.013*``won'' + 0.012*``team'' \\+ 0.011*``Olympics'' + 0.010*``Canada''}} \\
    & & \\
    & & \\
    \hline
    \multirow{3}{*}{5} & \multirow{3}{*}{The performance of female athletes} & \multirow{3}{*}{\makecell[c]{0.019*``women'' + 0.015*``final'' + 0.011*``Biles'' + 0.009*``Tokyo'' \\+ 0.009*``champion'' + 0.009*``British'' + 0.008*``gold'' + 0.006*``200m'' \\+ 0.006*``freestyle'' + 0.006*``100m''}} \\
    & & \\
    & & \\
    \hline
    \multirow{3}{*}{6} & \multirow{3}{*}{Competition rights and disputes} & \multirow{3}{*}{\makecell[c]{0.030*``athletes'' + 0.015*``Olympic'' + 0.013*``Paris'' + 0.013*``IOC'' \\+ 0.012*``women'' + 0.010*``Olympics'' + 0.009*``Games'' + 0.008*``Russian'' \\+ 0.007*``sports'' + 0.007*``country''}} \\
    & & \\
    & & \\
    \hline
    \end{tabular}%
    }
  \end{center}
  \label{tab2}%
\end{table*}%

\end{document}